\pdfoutput=1

\documentclass[11pt]{article}

\usepackage[]{acl}

\usepackage{times}
\usepackage{latexsym}

\usepackage[T1]{fontenc}

\usepackage[utf8]{inputenc}

\usepackage{microtype}

\usepackage{inconsolata}

\usepackage{graphicx}

\usepackage{xcolor,pifont}
\newcommand*\colourcheck[1]{%
  \expandafter\newcommand\csname #1check\endcsname{\textcolor{#1}{\ding{52}}}%
}
\colourcheck{blue}
\colourcheck{green}
\colourcheck{red}
\usepackage{graphicx}
\usepackage{amsmath}
\usepackage{amssymb}
\usepackage{booktabs}
\usepackage{dsfont}

\usepackage{amssymb}
\usepackage{adjustbox}
\usepackage{url}
\usepackage{colortbl}
\usepackage{multicol}

\usepackage{amsmath}
\usepackage{lipsum}  
\usepackage{adjustbox}
\usepackage{booktabs}
\usepackage{multicol}
\usepackage{multirow}

\usepackage{algorithm}
\usepackage{adjustbox}
\usepackage{mathtools}

\usepackage{enumitem}
\usepackage[noend]{algpseudocode}
\usepackage{graphicx}
\usepackage{subcaption}
\usepackage{bbm}
\usepackage{arydshln}

\algnewcommand\algorithmicforeach{{for each}}
\algdef{S}[FOR]{ForEach}[1]{\algorithmicforeach\ #1\ \algorithmicdo}

\usepackage{amsthm}

\usepackage{siunitx}

\title{\textsc{Melt}: Materials-aware Continued Pre-training \\for Language Model Adaptation to Materials Science}

\author{Junho Kim\textsuperscript{1}\thanks{\ \ These authors contributed equally to this work.}, 
  \textbf{Yeachan Kim\textsuperscript{1*}},
  \textbf{Jun-Hyung Park\textsuperscript{2}},
  \textbf{Yerim Oh\textsuperscript{1}},
  \textbf{Suho Kim\textsuperscript{1}},
  \textbf{SangKeun Lee\textsuperscript{1,3}} \\
  \textsuperscript{1}Department of Artificial Intelligence, Korea University, Seoul, Republic of Korea \\
  \textsuperscript{2}Division of Language \& AI, Hankuk University of Foreign Studies, Seoul, Republic of Korea \\
  \textsuperscript{3}Department of Computer Science and Engineering, Korea University, Seoul, Republic of Korea \\
  \texttt{\{monocrat, yeachan, yerim0210, suho6347, yalphy\}@korea.ac.kr, jhp@hufs.ac.kr} \\
}

\begin{document}
\maketitle

\begin{abstract}

We introduce a novel continued pre-training method, \textsc{Melt} (\textbf{M}at\textbf{e}ria\textbf{l}s-aware continued pre-\textbf{t}raining), specifically designed to efficiently adapt the pre-trained language models (PLMs) for materials science. Unlike previous adaptation strategies that solely focus on constructing domain-specific corpus,  \textsc{Melt} comprehensively considers both the corpus and the training strategy, given that materials science corpus has distinct characteristics from other domains. To this end, we first construct a comprehensive materials knowledge base from the scientific corpus by building semantic graphs. Leveraging this extracted knowledge, we integrate a curriculum into the adaptation process that begins with familiar and generalized concepts and progressively moves toward more specialized terms. We conduct extensive experiments across diverse benchmarks to verify the effectiveness and generality of \textsc{Melt}. A comprehensive evaluation convincingly supports the strength of \textsc{Melt}, demonstrating superior performance compared to existing continued pre-training methods. In-depth analysis of \textsc{Melt} also shows that \textsc{Melt} enables PLMs to effectively represent materials entities compared to the existing adaptation methods, thereby highlighting its broad applicability across a wide spectrum of materials science\footnote{Our code is available at \url{https://github.com/JunhoKim94/MELT}}.

\end{abstract}

\section{Introduction}

\begin{figure}[t]
\centering
\includegraphics[width=\linewidth]{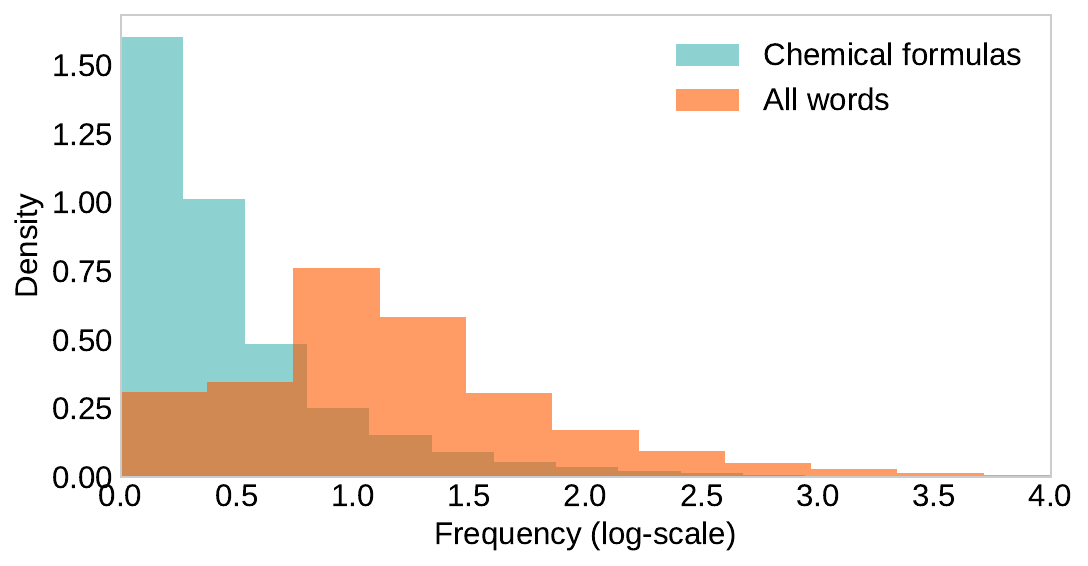}
\caption{Frequency histograms of all words and chemical formulas on materials science corpus (150K materials-related scientific papers).}
\label{fig:formula_frequency}
\end{figure}

Materials science encompasses interdisciplinary studies concerning the behaviors, properties, and applications of materials. Given the vast search space in materials science, deep learning-based approaches have emerged as significant avenues to accelerate the entire research pipeline \citep{mat2vec_nature19,matscholar,olivetti2020data}. Specifically, methods centered on natural language processing (NLP) have provided promising results across a number of materials tasks, such as materials entity recognition \citep{matscholar,sofc} and synthesis action retrieval \cite{wang2022ulsa}. However, the limited and scarce nature of datasets in this domain poses substantial challenges for developing models that generalize well across a broad range of materials entities \cite{matscinlp_acl23}.

One promising approach to addressing this limitation involves adapting the pre-trained language models (PLMs) for materials science by continuously pre-training them on materials science corpus \cite{matscibert,batterybert}. However, these methods have predominantly focused on constructing the domain-specific \textbf{\textit{corpora}} used in the continued pre-training process, neglecting the \textbf{\textit{training strategies}} employed in the adaptation process. This oversight can lead to inefficiencies in capturing domain-specific nuances and knowledge. For example, the chemical formulas (e.g., $\text{LiCoO}_{2}$), which are fundamental terms in the materials science field, are typically infrequent words (Figure \ref{fig:formula_frequency}). Therefore, these domain-specific terms are often inadequately captured by the random masking strategy used in previous studies, leading to sub-optimal adaptation results.

In this paper, we propose a novel continued pre-training method, namely \textsc{Melt} (\textbf{M}at\textbf{E}ria\textbf{L}s-aware continued pre-\textbf{T}raining) that learns the knowledge of materials science through the tailored masking strategy to materials science. To this end, we begin by extracting chemical entities from the materials science corpus. However, these entities alone do not encompass all the fundamental concepts of materials science, specifically the \textit{\textbf{structure-property-processing-performance}} paradigm \cite{william1989structure}. To bridge this gap, we expand the coverage of materials knowledge by constructing semantic graphs that integrate relevant concepts and missing entities. Building on this enhanced materials knowledge, we introduce curriculum into the adaptation process, starting with more familiar and interconnected entities and gradually advancing to more specialized and less common terms. This comprehensive adaptation allows PLMs to effectively learn materials science knowledge.

To verify the efficacy of \textsc{Melt}, we perform extensive experiments, spanning a wide range of downstream tasks for materials science on both generation and classification tasks. Comprehensive evaluation results clearly support the superiority of \textsc{Melt} over strong baselines. Moreover, we verify that \textsc{Melt} enables PLMs to effectively learn the materials by thoroughly transforming both the training corpora and adaptation strategies, underscoring the efficacy of the tailored approach to materials science. In summary, the contributions of this paper include the following:

\begin{itemize}
    \item We propose \textsc{Melt}, a novel continued pre-training method to adapt the pre-trained language models for materials science.
    
    \item We introduce a method for constructing a semantic graph of materials entities to widen the coverage of materials knowledge.
    
    \item We demonstrate that \textsc{Melt} substantially improves the performance compared to previous adaptation methods for materials science.
\end{itemize}

\section{Related Works}

\subsection{NLP in Materials Science}

The increasing number of textual datasets in materials science (e.g., scientific publications, patents) has facilitated the use of NLP-based approaches to address various materials tasks, such as relation classification \cite{mysore2019materials, mullick2024matscire} and materials entity extraction \cite{matscholar, sofc}. For example, \citet{matscholar} proposed a bidirectional LSTM tagger for named entity recognition on tags associated with the well-known materials science tetrahedron (i.e., structure, property, processing, performance). \citet{mat2vec_nature19} demonstrated promising results with embedding-based unsupervised methods for understanding chemistry knowledge and chemical properties. Beyond embedding models, \citet{matbert_patterns22} introduced PLMs trained on a materials science corpus following the BERT procedure \cite{bert}. Similarly, \citet{matscibert} and \citet{batterybert} adapted SciBERT \cite{scibert} and BERT \cite{bert} to the domains of general materials and battery-specific corpora by additionally training on the domain-specific corpus. Recently, HoneyBee \cite{honeybee} suggested the materials domain-specific instruction data to fine-tune the large language models.

Previous studies have demonstrated promising results in adapting PLMs to materials science. However, these methods have primarily focused on adapting the corpus for materials science while employing a basic random masking strategy \cite{bert}. Such a domain-agnostic approach potentially prevents the models from adequately learning about chemical entities and diverse formulas, which often fall into the less frequent tail distributions in word frequency. In contrast, \textsc{Melt} is a tailored approach to materials science, focusing on both the adaptation corpus and learning strategies to adapt PLMs for materials science.

\begin{figure*}[t]
\centering
\subfloat[]{%
  \includegraphics[width=\textwidth]{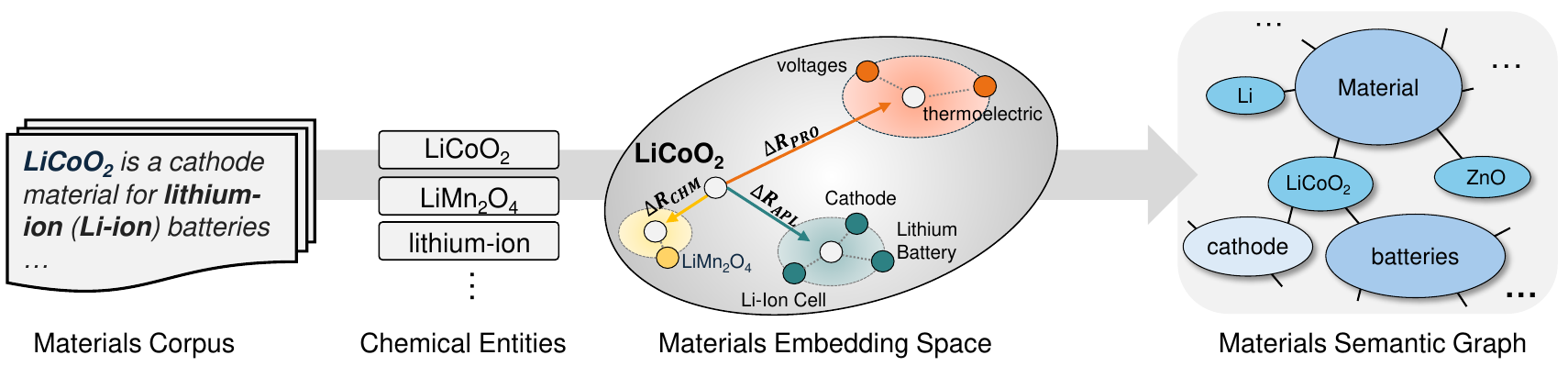}%
}

\subfloat[]{%
  \includegraphics[width=\textwidth]{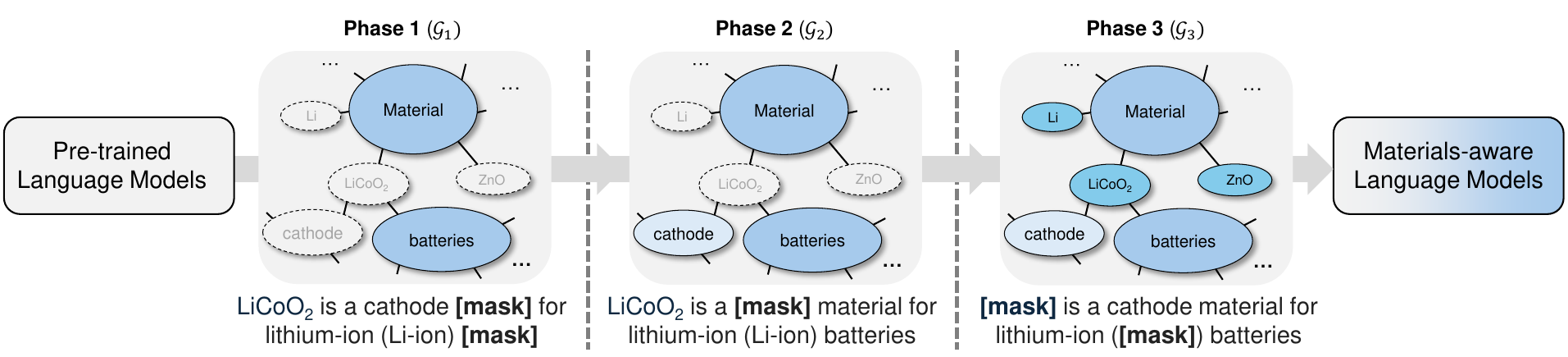}%
  \label{fig:curriculum} 
}

\caption{Overall adaptation process of PLMs with \textsc{Melt}. Starting from the materials science corpus, we extract chemical entities (e.g., chemical names and formulas). These entities are then expanded to related terms based on the materials embeddings with their compositional property, resulting in materials semantic graphs. Based on the constructed graphs, \textsc{Melt} performs curriculum masking over materials corpus (e.g., three phases in this example).}
\label{fig:main_model}
\end{figure*}

\subsection{Continued Pre-training of PLMs}
Building PLMs from scratch requires substantial computational resources; therefore, continued pre-training has garnered significant attention, especially in scientific domains. For example, \citet{baseline_dsp} adapted PLMs trained on general corpora to various domains (e.g., computer science, biomedical) by performing random-based masked language modeling on the target domain corpus. \citet{baseline_entity_bert} performed the domain adaptation by selectively masking entities based on the trained taggers. Subsequently, \citet{baseline_diff} proposed Diff-Masking that utilizes the frequency difference between generic and target domains to perform masked language modeling on domain-specific terms. In addition to the masking strategy, several works studied the catastrophic forgetting problem when adapting PLMs to other domains \cite{jin2022lifelong,baseline_das}. 

In contrast, \textsc{Melt} has distinct characteristics compared to the previous continued pre-training methods. First, previous works have designed a domain-agnostic masking strategy based on random sampling and word frequency. However, \textsc{Melt} involves a tailored masking strategy to materials science by comprehensively constructing materials knowledge based on the fundamental concepts of materials (i.e., structure, property, processing, performance). Second, \textsc{Melt} integrates a curriculum into the adaptation process, providing better local optima compared to naive adaptation methods \cite{bengio2009curriculum}.

\section{\mbox{Materials-aware Continued Pre-training}}

We elaborate on the proposed continued pre-training method, \textsc{MELT}. The key strategy is to extract the material entities from the scientific papers and inject the extracted knowledge of the materials into the PLMs through a masked language modeling (MLM) objective. To this end, we first extract the materials entities from the materials science corpus. From these entities, we expand the materials knowledge by semantically augmenting the related entities and properties. We then start pre-training on the basic materials entities and progressively move more specialized knowledge in a curriculum manner. The overall procedures are described in Figure \ref{fig:main_model}.

\subsection{Continued Pre-training through MLM}
We start by setting the training objective of continued pre-training. Following the promising results of the previous study \cite{baseline_dsp}, we adapt the PLMs to different domains by performing MLM \cite{bert} on the domain-specific corpus related to materials science. Specifically, it masks out a small portion of the input sequence (\textit{i.e.}, replace the original word with the special token [mask]) and trains the model to predict the original tokens. Formally, let the set of words to be masked be denoted as $\mathcal{G}$, the MLM training objective is as follows:
\begin{equation}
    \mathcal{L}(x, \hat{x};\theta) = \sum_{w_{m} \in \mathcal{G}, w_{m} \in x} \log P(y = w_{m}|\hat{x};\theta)
\end{equation}
where $x$ and $\hat{x}$ denote the original and masked input sequences based on the set of masking words $\mathcal{G}$, respectively. In a typical pre-training objective, the set $\mathcal{G}$ is randomly determined without considering the importance of words in the target domain.
However, such a random strategy makes the PLMs poorly learn the sparse domain-specialized terms \cite{baseline_diff}, as these terms are rarely selected for the training. Therefore, the objective of \textsc{Melt} is to fill the masking set $\mathcal{G}$ with the materials knowledge to adapt the PLMs to the domains of materials science effectively.

\subsection{Materials-aware Entity Masking}\label{sect:chem_masking}
\paragraph{Chemical Entity Extraction}\label{sect:chem_extract}

While the general domain corpus is widely spread across diverse sources and formats (e.g., Wikipedia, News, SNS), the large-scale corpus for materials science mainly exists in the form of scientific publications (e.g., patents, papers). Moreover, it has several distinct characteristics. First, the materials corpus includes a number of chemical formulas (e.g., H$_2$O,  LiCoO$_2$), which are crucial components to understanding materials science. 
For example, in materials benchmarks for token classification, such as named entity recognition, the formulas roughly take up 20\% among all entities\footnote{Based on the statistics of MatScholar and SOFC.}. 
Secondly, materials corpus frequently include various representations of domain-specific jargon and abbreviations (e.g., "THF" for \textit{Tetrahydrofuran}, "AZT" for \textit{3'-azido-3'-deoxythymidine}).

To extract these terms to learn materials knowledge, we follow the hybrid approach of ChemDataExtractor \cite{chemdataextractor,chemdataextractor2}, utilizing both dictionary-based mapping and CRF taggers to identify the materials entities. The extracted terms include not only elemental and compound names but also materials characteristics such as density, melting point, and electrical conductivity. This wealth of material-aware entity extraction provides a solid foundation for the continued pre-training.

\paragraph{Semantic Graph for Knowledge Expansion}

While the various types of material-aware entities from previous extraction phases involve knowledge about materials, they still lack the fundamental concept of materials science about {structure-property-processing-performance} paradigm within materials science \cite{william1989structure}.

To augment these fundamental relations, we construct semantic graphs of materials entities by connecting related terms and finding similar yet missing entities with the seed entities.
To this end, we first train a lightweight embedding model \cite{mat2vec_nature19} on the materials corpus \footnote{Details for word embeddings are in Appendix \ref{appendix:mat2vec}}. 
Based on the learned embeddings of materials terms and entities, we leverage the compositional property of the embedding representations \cite{word2vec} where the addition of the two embeddings can indicate the compositional meaning (e.g., the sum of the two embeddings, \textit{Vietnam} and \textit{Capital}, can infer the meaning of \textit{Hanoi}). 
For example, to find out the \textit{Application} of the \textit{carbon} entity, adding the embedding about \textit{Application} to the embedding of \textit{carbon} can infer new entities related to the property of carbon (Detailed example can be found in Appendix \ref{appendix:expansion_example}). 

To obtain the pivot vectors representing fundamental concepts of materials science, we extract the example pairs of each concept\footnote{We consider six fundamental concepts: Material, Property, Application, Characterization, Descriptor, Symmetry/Phase. The details for each relation are listed in Appendix \ref{appendix:relation}.} from MatKG \cite{matkg} and build the concept embeddings by subtracting the word embeddings of the pair of each concept. Formally, let the set of the word pairs for the concept $R$ be denoted as $\mathcal{S}_{R}$, and the embedding for the concept $R$ is derived as follows:
\begin{equation}\label{eq:relation_subtract}
    e(R) = \frac{1}{|\mathcal{S}_{R}|} \sum_{(w_{a}, w_{b}) \in \mathcal{S}_{R}}(e(w_{a}) - e(w_{b}))
\end{equation}
where $w_{a}$ and $w_{b}$ are the subject and object for the relation concept $R$, and $e(w)$ denotes the trained embedding for the word $w$. Based on the embeddings of each relation concept, we build the semantic graph where nodes represent entities and edges represent similarity between entities. Specifically, we make the connections based on the cosine similarity between all words in trained embeddings and compositional embedding of extracted entities and concepts as follows:
\begin{equation}\label{eq:graph_sim}
    \frac{(e(w_i) + e(R)) \cdot e(\mathcal{V})}{\lVert(e(w_i) + e(R))\rVert \lVert e(\mathcal{V}) \rVert}, \forall w_i \in \mathcal{G}
\end{equation}
where $e(\mathcal{V})$ denotes the embeddings for all words. However, all words connected by this procedure are not closely related to each entity. Therefore, we only leave the top-k elements based on their similarity. Throughout this process, we obtain a richer set of materials entities for MLM. The number of unique entities is in Appendix \ref{appendix:unique}

\begin{algorithm}[t]
\centering
\caption{Curriculum-based Entity Learning}\label{alg:curriculum}
\begin{flushleft}
\textbf{Input:} Node degree-based curriculum $\{\mathcal{G}_1 ... \mathcal{G}_K\}$, PLM parameters $\theta$, training corpus $D$, maximum training step $\tau$, masking probability $p_m$, 

\begin{algorithmic}[1]
\State \textbf{for} token sequence $T$ in $D$ \textbf{do}
\State \ \ \ Search entities $e$ in $T$.
\State \ \ \ Order entities according to $\{\mathcal{G}_1 ... \mathcal{G}_K\}$
\State \textbf{end for}
\State \textbf{while} step $<$ $\tau$ \textbf{do}
\State \ \ \ \textbf{for} stage $i=1,2,...,K$ \textbf{do}
\State \ \ \ \ \ \ $T'$ $\leftarrow$ MaskingEntities($T$, $\mathcal{G}_i$, $p_m$)
\State \ \ \ \ \ \ Train with MLM for PLM $\theta$ on $T'$
\State \ \ \ \textbf{end for} 
\State \textbf{end while} 
\State \textbf{return} PLM parameters $\theta$
\end{algorithmic}
\end{flushleft}
\label{alg:melt}
\end{algorithm}

\subsection{Curriculum-based Entity Learning}
While continued pre-training with the tailored masking can effectively train the knowledge of materials sciences, learning with specialized entities from the beginning makes the training unstable, leading to poor generalization performance \cite{pointer_cl_acl2019,meta_curricular_aaai2021,CCM_emnlp2022}. Inspired by the human learning process, which learns easy concepts first and moves to harder ones, we introduce a curriculum into the adaptation process.

To define the difficulty within the set of extracted materials terms $\mathcal{G}$, we utilize the node degree\footnote{Node degree is the number of connected edges to nodes.} in the constructed semantic graph in section \ref{sect:chem_masking}. Since the more connected entities are likely to be more familiar and well-defined within the corpus, they can provide a straightforward criterion for prioritizing the learning sequence. Therefore, we gradually mask materials entities from those with higher node degrees (i.e., fundamental and common concepts) to those with lower node degrees (i.e., niche and specialized concepts).

We thus calculate the node degree of all entities and decompose the set of entities $\mathcal{G}$ into a stratified format, i.e., $\mathcal{G}$ = \{$\mathcal{N}_{1}, \mathcal{N}_{2}, ..., \mathcal{N}_{K}\}$ where $K$ is the number of stages for the curriculum, and $\mathcal{N}_{1}$ consists of the easiest entities while $\mathcal{N}_{K}$ includes the hardest entities. 
Then, we gradually expand the materials entities from the easiest entity set $\mathcal{G}_{1}$ to the hardest entity set to construct the curriculum set $\mathcal{G}_{i}$ as follows:
\begin{equation}
    \mathcal{G}_{i} = \mathcal{G}_{i-1} \cup \mathcal{N}_{i}, \ \ (\mathcal{G}_{1} = \mathcal{N}_{1})
\end{equation}
By utilizing the curriculum sets, we progressively mask the entities connected to previously learned entities for each stage.
The detailed process is represented in Algorithm \ref{alg:melt}. 

\begin{table*}[t]
\centering
\caption{Evaluation results on MatSci-NLP. \textbf{NER}: Named-Entity Recognition, \textbf{RC}: Relation Classification, \textbf{EAE}: Event Argument Extraction, \textbf{PC}: Paragraph Classification, \textbf{SAR}: Synthesis Action Retrieval, \textbf{SC}: Sentence Classification, \textbf{SF}: Slot Filling. The best and the second best results are highlighted in \textbf{boldface} and \underline{underline}, respectively.}
\label{tab:MatSci-NLP_results}
{%
\begin{adjustbox}{width=\linewidth}
\begin{tabular}{cccccccccc}
\toprule
\multirow{2}{*}{Method}                                 & \multirow{2}{*}{Metric}       & \multicolumn{7}{c}{MatSci-NLP}          &                                                      \\  

 &  & NER & RC & EAE & PC & SAR & SC & SF & Overall \\ \midrule
\multirow{2}{*}{\begin{tabular}[c]{@{}c@{}}SciBERT\\ \cite{scibert}\end{tabular}} & Micro-F1 & 0.738 & 0.818 & 0.458 & 0.671 & 0.701 & 0.909 & 0.500 & 0.693 \\
 & Macro-F1 & 0.517 & 0.600 & 0.290 & 0.568 & 0.528 & 0.612 & 0.258 & 0.482 \\ \midrule \midrule
 
 \multirow{2}{*}{\begin{tabular}[c]{@{}c@{}}DSP\\ \cite{baseline_dsp}\end{tabular}} & Micro-F1 & 0.733 & 0.857 & 0.483 & 0.704 & 0.772 & \textbf{0.914} & 0.568 & 0.722 \\
 & Macro-F1 & 0.500 & \underline{0.631} & \underline{0.313} & 0.579 & 0.649 & \textbf{0.630} & 0.322 & 0.518 \\ \midrule
 
  \multirow{2}{*}{\begin{tabular}[c]{@{}c@{}}EntityBERT\\ \cite{baseline_entity_bert}\end{tabular}} & Micro-F1 & 0.728 & 0.805 & \underline{0.495} & \textbf{0.765} & 0.739 & 0.897 & 0.559 & 0.713 \\
 & Macro-F1 & 0.528 & 0.567 & 0.311 & \textbf{0.699} & 0.621 & 0.581 & 0.303 & 0.516 \\ \midrule
 
\multirow{2}{*}{\begin{tabular}[c]{@{}c@{}}DAS\\ \cite{baseline_das}\end{tabular}} & Micro-F1 & 0.770 & 0.848 & 0.478 & 0.672 & \underline{0.778} & 0.902 & \underline{0.592} & \underline{0.725} \\
 & Macro-F1 & 0.567 & 0.628 & 0.292 & 0.607 & 0.641 & 0.607 & 0.356 & 0.528 \\ \midrule
 
\multirow{2}{*}{\begin{tabular}[c]{@{}c@{}}Diff-Masking\\ \cite{baseline_diff}\end{tabular}} & Micro-F1 & \underline{0.771} & \underline{0.858} & 0.471 & 0.686 & 0.769 & 0.879 & 0.589 & 0.721 \\
 & Macro-F1 & \underline{0.575} & \textbf{0.641} & 0.301 & 0.573 & \textbf{0.687} & \underline{0.622} & \underline{0.373} & \underline{0.539} \\ \midrule
 


\cellcolor[gray]{.9}  & \cellcolor[gray]{.9}Micro-F1 & \cellcolor[gray]{.9}\textbf{0.786} & \cellcolor[gray]{.9}\textbf{0.860} & \cellcolor[gray]{.9}\textbf{0.498} & \cellcolor[gray]{.9}\underline{0.728} & \cellcolor[gray]{.9}\textbf{0.798} & \cellcolor[gray]{.9}\underline{0.911} & \cellcolor[gray]{.9}\textbf{0.610} & \cellcolor[gray]{.9}\textbf{0.741} \\

\cellcolor[gray]{.9}\multirow{-2}{*}{\textsc{Melt} (ours)} & \cellcolor[gray]{.9}Macro-F1 & \cellcolor[gray]{.9}\textbf{0.593} & \cellcolor[gray]{.9}0.620 & \cellcolor[gray]{.9}\textbf{0.341} & \cellcolor[gray]{.9}\underline{0.647} & \cellcolor[gray]{.9}\underline{0.685} & \cellcolor[gray]{.9}{0.613} & \cellcolor[gray]{.9}\textbf{0.395} & \cellcolor[gray]{.9}\textbf{0.556} \\
\bottomrule
\end{tabular}%
\end{adjustbox}
}
\end{table*}

\section{Experiments}
In this section, we verify the efficacy of the \textsc{Melt}. Specifically, we answer the following four questions through extensive experiments and analysis:
\begin{itemize}
    \item[\textbf{Q1}]\textbf{(Adaptability) } Does \textsc{Melt} enable better adaptation than existing methods across diverse benchmarks? (\S\ref{sec:main}, \S\ref{sec:curriculum} \S\ref{sec:generation})
    \item[\textbf{Q2}]\textbf{(Transferability) } How the extracted materials knowledge from \textsc{Melt} is related and affect to downstream tasks? (\S\ref{sec:ablation}, \S\ref{sect:relevance}, \S\ref{sect:entity_centric})
    \item[\textbf{Q3}]\textbf{(Efficiency) } Does \textsc{Melt} offer better efficiency than existing continued pre-training methods for domain adaptation? (\S\ref{sec:efficiency})
    \item[\textbf{Q4}]\textbf{(Insights) } What materials science knowledge is extracted and learned during the continued pre-training? (\S\ref{sec:masking_words})
\end{itemize}

\subsection{Experimental Setups}

\paragraph{Baselines} 
To confirm the effectiveness of \textsc{Melt}, we mainly compare ours with strong domain adaptation methods for PLMs: \textbf{DSP} \cite{baseline_dsp}, \textbf{EntityBERT} \cite{entitybert_bionlp21}, \textbf{Diff-Masking} \cite{baseline_diff}, and \textbf{DAS} \cite{baseline_das}. The detailed experimental setups for baselines are described in Appendix \ref{appendix:baselines}. 


\paragraph{Pre-training} To adapt the PLMs to the domain of materials science, we leverage the 150K scientific papers related to materials science following the previous work \cite{matscibert}. In \textsc{Melt}, we implement iterative curriculum learning, consisting of 10K steps for warm-up training and 10K steps for each subsequent curriculum stage. We set the number of curriculum stages ($K$) to 3 based on the empirical analysis. Detailed analysis is represented in Appendix \ref{appendix:curriculum_num}.


\paragraph{Downstream Tasks and Datasets} To demonstrate the diverse aspects of the \textsc{Melt}, we compare the models on both generation and classification tasks.
For generation tasks, we evaluate each baseline using the MatSci-NLP dataset \cite{matscinlp_acl23}, which comprises seven materials-related tasks (e.g., materials entity recognition, slot filling). Detailed information on each task within MatSci-NLP can be found in Appendix \ref{appendix:MatSci-NLP}. For classification tasks, we adopt four different tasks following the previous research \cite{matscibert}, which include NER (MatScholar \cite{matscholar}, SOFC-EXP \cite{sofc}), paragraph classification (Glass Science \cite{glass}), and slot filling (SOFC-Filling \cite{sofc}) tasks. Detailed evaluation metrics are shown in Appendix \ref{appendix:evaluation}.


\paragraph{Backbones}
Following prior work \cite{matscibert}, we apply each adaptation method to the SciBERT \cite{scibert}, which is an encoder-based model pre-trained on a scientific corpus comprising approximately 1.14 million documents. 
For generation tasks, following the recent setup from the previous work \cite{matscinlp_acl23}, we construct an encoder-decoder transformer model by integrating a transformer decoder with the encoder model, which pre-trained on each baseline.



\begin{table}[t]
\centering
\caption{Ablation results on MatSci-NLP. CEL and MEM indicate curriculum-based entity learning and materials-aware entity masking, respectively. The values in parentheses represent the performance difference to \textsc{Melt}. Detailed results are shown in Appendix \ref{appendix:ablation}.}
\label{tab:ablation}
\begin{tabular}{@{}lcc@{}}
\toprule
\multirow{2}{*}{Method} & \multicolumn{2}{c}{MatSci-NLP (Overall)} \\ \cmidrule(l){2-3} 
 & Micro-F1 & Macro-F1 \\ \midrule
 \cellcolor[gray]{.9}\textsc{Melt} (ours) & \cellcolor[gray]{.9}{0.741} & \cellcolor[gray]{.9}{0.556} \\ \midrule
w/o CEL & 0.729 {\small (-0.012)} & 0.542 {\small (-0.014)}\\ 
w/o CEL, MEM & 0.713 {\small (-0.028)}& 0.516 {\small (-0.040)}\\
 \bottomrule
\end{tabular}%
\end{table}

\subsection{Main Results}\label{sec:main}
Table \ref{tab:MatSci-NLP_results} presents the overall performance results on the MatSci-NLP benchmark. The comparison indicates that \textsc{Melt} yields the best adaptation outcomes compared to previous methods. Specifically, \textsc{Melt} enhances the performance of the backbone model by an average of 6.9\% and 15.3\% in terms of Micro-F1 and Macro-F1 scores, respectively, across all tasks. This indicates the broad applicability of the proposed method to various materials science tasks. It is also noteworthy that domain-agnostic methods (i.e., Diff-Masking, DAS) reveal limited performance improvement compared to the random masking baseline (i.e., DSP \cite{baseline_dsp}). It underscores the effectiveness of the domain-specific approach in continued pre-training. Overall, the results strongly support the superiority of the proposed method in adapting PLMs to materials science.

\subsection{Ablation Study}\label{sec:ablation}
We perform ablation studies to confirm whether the components in \textsc{Melt} are crucial in producing better PLMs for materials science. Table \ref{tab:ablation} shows the ablation results about two components: (i) Materials-aware entity masking (MEM) by building a semantic graph of chemical entities\footnote{Note that models without the expansion do not have curriculum components as well because the proposed curriculum learning depends on the results of the expansion procedure.} (ii) Curriculum-based entity learning (CEL) based on the node degree of the constructed graph. Notably, we find that omitting each component from \textsc{Melt} leads to a substantial drop in performance, providing empirical verification of the proposed method in continued pre-training. Specifically, we highlight the results of the model without both CEL and MEM in \textsc{Melt}, which uses only the chemical entities similar to those in EntityBERT. Although this model exhibits inferior performance compared to the random masking baseline, it achieves significantly improved performance when expanding its knowledge with fundamental concepts from the materials science tetrahedron and learning it through a well-defined curriculum.

\begin{table}[t]
\centering
\caption{Comparison of different curriculum strategies with \textsc{Melt}. The best results are highlighted in \textbf{boldface}. }
\label{tab:curriculum}
\begin{tabular}{@{}lcc@{}}
\toprule
\multirow{2}{*}{Curriculum Strategy} & \multicolumn{2}{c}{MatSci-NLP (Overall)} \\ \cmidrule(l){2-3} 
 & Micro-F1 & Macro-F1 \\ \midrule
No Curriculum & 0.729 & 0.542 \\ \hdashline
Frequency & 0.736 & 0.545 \\ 
Concept & 0.736 & 0.551 \\ 
Masking Ratio & 0.737 & 0.547 \\
Reverse & 0.730 & {0.552} \\ \midrule
 \cellcolor[gray]{.9}\textsc{Melt} (ours) & \cellcolor[gray]{.9}\textbf{0.741} & \cellcolor[gray]{.9}\textbf{0.556} \\ \bottomrule
\end{tabular}%
\end{table}

\begin{table}[t]
\centering
\caption{Evaluation results on four classification tasks. The best and second best results are highlighted in \textbf{boldface} and \underline{underline}, respectively.}
\begin{adjustbox}{width=\linewidth}
\begin{tabular}{@{}lcccccccc@{}}
\toprule
Methods & NER$_{\text{SOFC}}$ & NER$_{\text{MS}}$ & PC & SF \\ \midrule
SciBERT  & 80.3  & 85.3    & 94.2  & 59.1     \\ \hdashline
DSP   & 80.4   & \underline{85.8}  & \underline{95.2}  & 61.4      \\
EntityBERT  & 80.0 & 85.1 & 94.7 & \underline{62.2}       \\
DAS  & 79.7 & 85.5 & 95.1 & 58.4     \\ 
Diff-Masking  & \underline{80.6} & 85.6  & 94.7 & 62.0     \\ \midrule
\cellcolor[gray]{.9}\textsc{Melt} (ours)              & 
 \cellcolor[gray]{.9}\textbf{81.1}  & \cellcolor[gray]{.9}\textbf{86.0}   & \cellcolor[gray]{.9}\textbf{95.7}  & \cellcolor[gray]{.9}\textbf{62.9}      \\ \bottomrule
\end{tabular}
\end{adjustbox}
\label{tab:main}
\end{table}

\subsection{Effect of the Curriculum Adaptation}\label{sec:curriculum}
To further investigate the effectiveness of the proposed curriculum learning, we compare ours with the existing curriculum-based MLM strategies: Frequency, Concept, Masking Ratio, and Reverse\footnote{Details for each strategy is described in the Appendix \ref{appendix:curriculum}}. Table \ref{tab:curriculum} presents the comparison results on the MatSci-NLP benchmark. We observe that the curriculum approach in \textsc{Melt} demonstrates superior performance compared to existing curriculum strategies. These results underscore the efficacy of considering node degree in materials semantic graphs when adapting PLMs for materials science.


\begin{figure}[t]
\centering
\includegraphics[width=\linewidth]{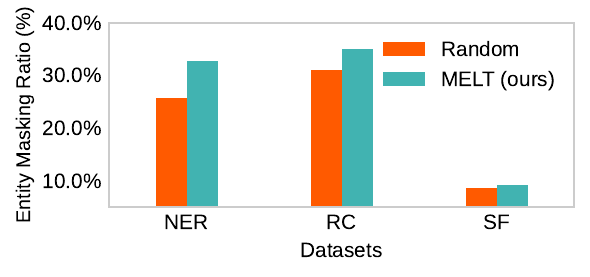}
\caption{Masking ratio (\%) of the materials entities in token classification datasets (NER, RC, SF).}
\label{fig:analysis_overlap}
\end{figure}

\begin{figure}[t]
\centering
\includegraphics[width=0.9\linewidth]{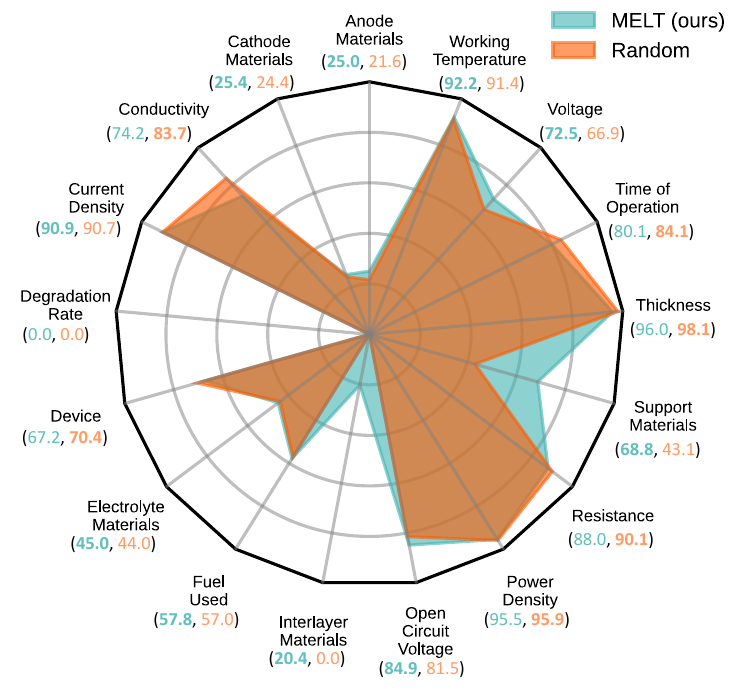}
\caption{Categorical accuracy (\%) for the classes of the chemical entities in the SF task.}
\label{fig:analysis_category}
\end{figure}

\subsection{Evaluation on Classification Tasks} \label{sec:generation}

To assess the generality of our model, \textsc{Melt}, we evaluate each baseline on four classification tasks following the settings from prior work \cite{matscibert}. We present the evaluation results on test sets in Table \ref{tab:main}. Similar to the results in generation tasks, \textsc{Melt} outperforms other baselines in most cases. These results demonstrate that PLMs adapted by \textsc{Melt} perform effectively across various classification tasks.

\subsection{Masking Relevance to Downstream Tasks}\label{sect:relevance}
We additionally analyze how the extracted knowledge relates to the entities of the downstream tasks. Here, we compare the proposed method with the random masking. For evaluation, we aggregate all words with their tags in NER (MatScholar and SOFC), RC, and SF tasks and calculate the overlapped ratio of entity tags (labels starting with B- or I-). Figure \ref{fig:analysis_overlap} shows the comparison results with the baseline. The results show that the proposed method, \textsc{Melt}, shows a significantly larger ratio of entity tags. It can be interpreted as our masking strategy successfully extracting domain-specific entities (e.g., chemical formulas, property).


\subsection{Effect of Materials-aware Entity Masking}\label{sect:entity_centric}
The extracted knowledge from \textsc{Melt} is centered on chemical entities (the first step of the proposed method), and the benchmark datasets for materials science involve a number of chemical entities. To verify the effectiveness of the entity-centric approach, we further analyze the performance of the fine-tuned model for the classes of the chemical entities (i.e., chemical names or formulas) in the slot-filling task (SOFC-Filling). Figure \ref{fig:analysis_category} shows the categorical performance of the entity-based (ours) and random-based approaches. We observe that our \textsc{Melt} achieves superior performance in all materials categories. Specifically, \textsc{Melt} outperforms random-based masking about 25\% on the Support materials class. Moreover, our \textsc{Melt} performs better than random masking across a variety of properties. These results indicate that materials-aware entity masking can improve the generalization ability of PLMs by learning from diverse material entities.

\begin{figure}[t]
\subfloat[NER$_{\text{SOFC}}$]{%
  \includegraphics[width=\linewidth]{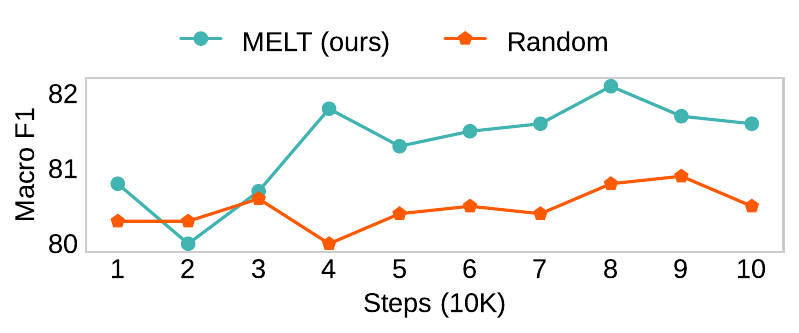}%
}

\subfloat[SF]{%
  \includegraphics[width=\linewidth]{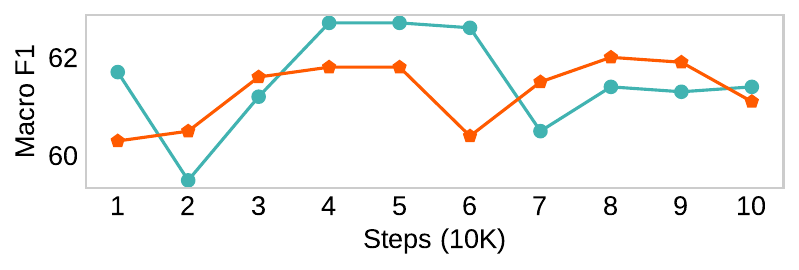}%
}

\caption{Comparison of the Macro F1 score for the SOFC-NER and SOFC-Filling test sets over the different number of pre-training steps.}
\label{fig:efficient}
\end{figure}

\subsection{Efficiency of Continued Pre-training}\label{sec:efficiency}
We analyze the training efficiency of our proposed \textsc{Melt} by comparing the performance of SOFC-NER and SOFC-Filling test sets per pre-training step with the baseline. The results are shown in Figure \ref{fig:efficient}. We observe that our \textsc{Melt} consistently outperforms the baselines after 40K steps pre-trained on both datasets. These results indicate that our materials-aware masking and curriculum learning strategies efficiently adapt the PLMs to materials science domains.

\begin{figure}[t]
\centering
 \includegraphics[width=\linewidth]{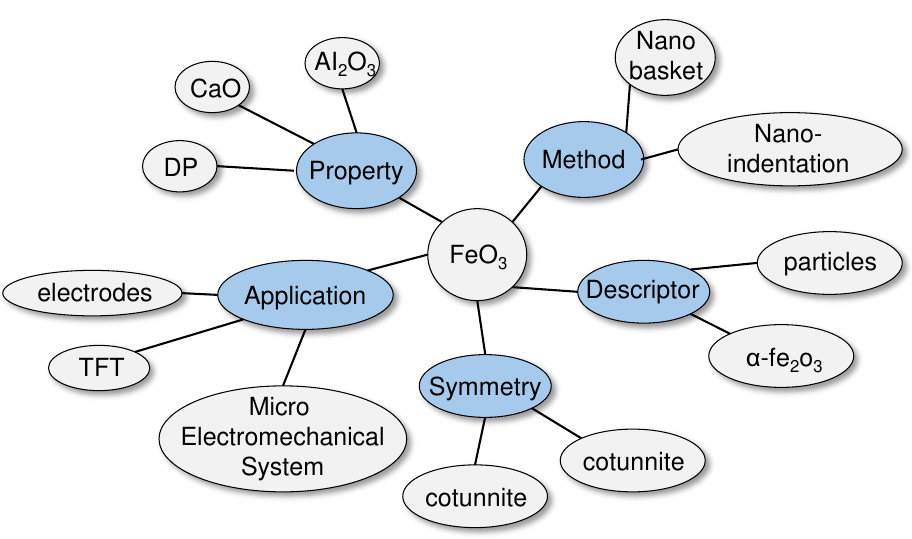}%
\caption{Example of the constructed graphs starting from the initially extracted chemical formula (FeO$_3$). Only the entities in white circles are used to construct the set of masking words $\mathcal{G}$.}
\label{fig:analysis_graph}
\end{figure}

\subsection{Extracted Materials Knowledge}\label{sec:masking_words}
Our \textsc{Melt} extracts chemical entities from the scientific corpus and expands the knowledge through a materials semantic graph. To validate the constructed knowledge, we sample a part of the graph with the five relation types (i.e., Property, Method, Application, Symmetry label, Descriptor). 
Figure \ref{fig:analysis_graph} represents the expansion of material entities (e.g., DP, electrodes, cotunnite) with five relation types from extracted chemical entities in corpora (i.e., FeO3)\footnote{Note that we only represent the additional entities that overlap with MatKG \cite{matkg}}. 
We observe that most extracted entities (e.g., electrodes, Micro Electromechanical Systems) represent the related concepts well, indicating the effectiveness of our automatic knowledge expansion. Moreover, even in cases of entities with incorrect relations (e.g., CaO, Al2O3 with the property relation), these entities are still materials-related words, demonstrating that such entities can be effectively used in the training stage. The overall results highlight the usefulness of the extracted knowledge from \textsc{Melt}. More examples can be found in Appendix \ref{appendix:extracted_example}

\section{Conclusion}
In this paper, we have proposed \textsc{{Melt}}, a novel continued pre-training method to adapt PLMs for materials science. Unlike previous approaches that typically focus on the adaptation corpus, our method comprehensively considers both the corpus and the training strategy in the adaptation process. Specifically, we have constructed the materials knowledge base by considering materials science tetrahedron \cite{william1989structure}. This extracted knowledge is then transferred to the PLMs in a structured, curriculum-based manner. We have conducted extensive experiments across diverse materials science benchmarks. The evaluation results convincingly demonstrate that our tailored approach yields superior performance on downstream tasks compared to domain-agnostic methods. In-depth analysis focusing on generality, efficiency, and insights has further supported the efficacy of \textsc{Melt} in continued pre-training, highlighting its broad value in the field of materials science.


\section*{Limitations}
While we have demonstrated that \textsc{Melt} effectively adapts the PLMs on materials science domains, there are several limitations that present valuable opportunities for future research. 
\paragraph{Further Analysis on Generation Tasks} 

We have mainly focused on improving the efficacy of continued pre-training on information retrieval tasks (e.g., NER, Slot Filling), aligning with previous works \cite{matscinlp_acl23, matbert_patterns22, matscibert}. However, the applicability of \textsc{Melt} on further generation tasks for material discovery, such as hypothesis or code generation \cite{codegen}, remains under-explored in this work. Nevertheless, considering the significant performance improvements of \textsc{Melt} achieved on generation benchmarks, MatSci-NLP, we believe that \textsc{Melt} is expected to work well within other generation tasks. We leave the exploration of this direction as promising future research.


\paragraph{Tokenization for Materials Sciences} 
We have expanded materials entities and masked them to adapt the PLMs on the materials domain in this work. During this process, we have observed that complex materials terms such as chemical formulas and substances split into multiple subwords (e.g., LiMnO$_2$ tokenized to \{LiMn, \#\#O, \#\#2\}). Such over-tokenization may limit the PLMs to learning meaningful representations in the pre-training stage, which leads to sub-optimal results on the downstream tasks \cite{opportune}. These observations highlight the necessity of materials domain-specific tokenizers for processing complex chemical terms, and we expect that future work in this direction will be another promising avenue for domain adaptation for PLMs.

\section*{Acknowledgements}
This work was supported by the National Research Foundation of Korea (NRF) grant funded by the Korea government (MSIT) (No.RS-2024-00415812 and No.2021R1A2C3010430) and Institute of Information \& communications Technology Planning \& Evaluation (IITP) grant funded by the Korea government (MSIT) (No.RS-2024-00439328, Karma: Towards Knowledge Augmentation for Complex Reasoning (SW Starlab), No.RS-2024-00457882, AI Research Hub Project, and No.RS-2019-II190079, Artificial Intelligence Graduate School Program (Korea University)).

\clearpage
\newpage
\newpage

\appendix

\begin{center}
\LARGE
\textbf{Appendix}    
\end{center}
\label{sec:appendix}

\section{Details for Entity Expansion}
\subsection{Example of Chemical Entity Expansion}\label{appendix:expansion_example}
\begin{figure}[h]
\centering
\includegraphics[width=\linewidth]{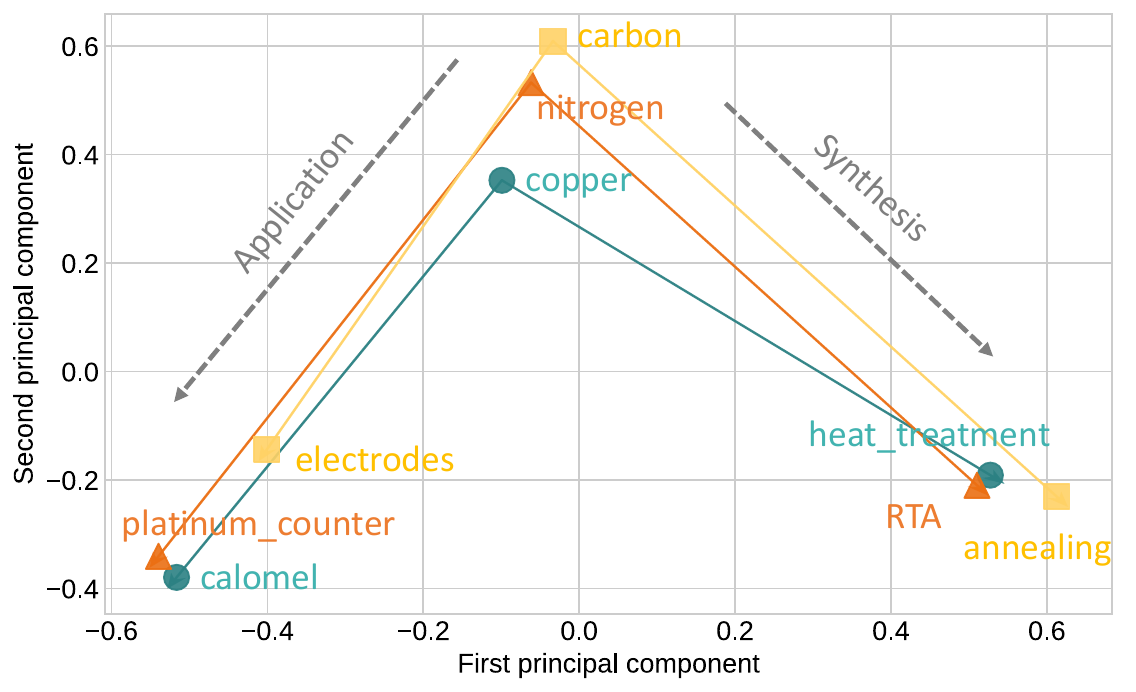}
\caption{2D Projection of the materials entities ({\textit{carbon, nitrogen, copper}}) having the relations of Applications (\textit{electrodes, platinum\_counter, calomel}) and Synthesis method (\textit{annealing, heat\_treatment, RTA}).}
\label{fig:method}
\end{figure}

\subsection{Details for Word Embedding Models} \label{appendix:mat2vec}

\begin{table}[h]
\centering
\caption{Hyper-parameter settings for our word embedding models}
\begin{tabular}{@{}lc@{}}
\toprule
Hyper-parameters & Values \\ \midrule
Vocabulary size & 876,911 \\
Epochs & 30 \\
Embedding sizes & 200 \\
Learning rate & 0.01 \\
Context Window & 8 \\
Sub-sampling threshold & 1e-4 \\
Number of negative samples & 15 \\ \bottomrule
\end{tabular}
\label{table:mat2vec}
\end{table}

Since the word embedding has been known to capture the relations between words or phrases \cite{word2vec}, we adopt the materials word2vec model, Mat2Vec \cite{mat2vec_nature19}, to expand the materials-related entities and their relations (Section 3.2). In this section, we detail the implementations for word embedding. Following the previous work \cite{mat2vec_nature19}, we tokenized our pre-training corpus, utilizing ChemDataExtractor \cite{chemdataextractor2} and their pre-defined rule-based tokenizer to construct the vocabulary. The vocabulary consisted of all words that occurred more than five times except chemical formulae. We leverage the skip-gram with negative sampling loss. The detailed hyperparameters are shown in Table \ref{table:mat2vec}

\subsection{Expansion Categories of Materials} \label{appendix:relation}
We consider six fundamental relations between materials entities, following the relations on the large-scale materials graph, MatKG \cite{matkg}. Specifically, they categorized text tokens into one of the following seven classes: Material, Property, Application, Synthesis Method, Characterization Method, Descriptor, and Symmetry/Phase Label. Regarding the traditional structure-property-processing-application paradigm in materials science \cite{william1989structure}, these entities encapsulate the total knowledge of any given concept. Thus, we expand the materials properties by utilizing six different categories, which are provided in MatKG \cite{matkg}.

\subsection{Detailed Statistics for Semantic Graph}
\label{appendix:unique}
\begin{table}[h]
\centering
\begin{tabular}{@{}lc@{}}
\toprule
 & \#Unique Entities \\ \midrule
\textsc{Melt} & 713,241 \\
\textsc{Melt} w/o expansion & 332,724 \\ \bottomrule
\end{tabular}
\caption{The number of Unique entities}
\label{tab:entitynum}
\end{table}
We construct the materials-aware semantic graph for the tailored masking strategy to materials science (Section 3.2). In this section, we detail the statistics for our semantic graph. Specifically, we compare the number of unique materials-aware entities before and after expansion. The details are shown in Table \ref{tab:entitynum}. The results indicate that our expansion methods provide more diverse material-aware entities (e.g., chemical properties and applications).

\section{Details for MatSci-NLP} 
\label{appendix:MatSci-NLP}

\begin{table}[h]
\centering
\caption{The detailed size and meta-datasets for MatSci-NLP \cite{matscinlp_acl23}.}
\begin{tabular}{@{}ccc@{}}
\toprule
Task & \multicolumn{1}{c}{\begin{tabular}[c]{@{}c@{}}Size\\ (\#Samples)\end{tabular}} & \multicolumn{1}{c}{\begin{tabular}[c]{@{}c@{}}Meta-Dataset\\ (\#Datasets)\end{tabular}} \\ \midrule
NER & 112,191 & 4 \\
RC & 25,674 & 3 \\
EAE & 6,566 & 2 \\
PC & 1,500 & 1 \\
SAR & 5,547 & 1 \\
SC & 9,466 & 1 \\
SF & 8,253 & 1 \\ \bottomrule
\end{tabular}
\label{tab:MatSci-NLP}
\end{table}

We compare our models on the multitask generation benchmark, MatSci-NLP \cite{matscinlp_acl23}, which includes seven different types of tasks with 13 datasets.
Specifically, MatSci-NLP consists of named-entity recognition (NER) \cite{sofc, matscholar, ner3, ner4}, relation classification (RC) \cite{ner3, ner4, rc}, event argument extraction (EAE) \cite{ner3, ner4}, paragraph classification (PC) \cite{glass}, synthesis action retrieval (SAR) \cite{sar}, sentence classification (SC) \cite{sofc}, and slot filling (SF) \cite{sofc}. 
The detailed statistics are represented in Table \ref{tab:MatSci-NLP}. 

\section{Implementation Details and Setups}

\subsection{Continued Pre-training Baselines} \label{appendix:baselines}
We compare our \textsc{Melt} with four strong continued pre-training methods for domain adaptation. For all baselines and \textsc{Melt}, we mask 15\% of the total tokens except Diff-Masking. We detail the implementations as follows:
\paragraph{DSP \cite{baseline_dsp}} is a naive approach to continually pre-train the PLMs for domain adaption by utilizing a random masking strategy. We consider the word-level approaches in our experiments following the masking strategies of MatSciBERT \cite{matscibert}.
\paragraph{EntityBERT \cite{entitybert_bionlp21}} is a method for masking tokens based on whether they are part of domain-specific entities by utilizing the NER model. The original paper leverages the PubMedBERT model, which was trained originally in the clinical domain. In this work, we utilize the ChemDataExtractor \cite{chemdataextractor2} to identify materials entities for masking.
\paragraph{Diff-Masking \cite{baseline_diff}} chooses tokens to mask based on anchor words, which are not commonly found in general domains $X_{PT}$ but commonly found in specific domains $X_T$. We leverage the $X_{PT}$ as the SciBERT pre-training corpus \cite{scibert} and $X_{T}$ as the corpus, which we constructed in section 3.2. Following the original paper, we mask 25\% of the total tokens and use 20 anchor words.
\paragraph{DAS \cite{baseline_das}} is an approach that adaptively updates the parameters by considering the importance of target domains. Following the original paper, we use the weights of contrastive learning $\lambda$ and temperature $\tau$ as 1.0 and 0.05, respectively.

\subsection{Hyper-parameters} \label{appendix:hyperparameter}
\paragraph{Pre-train.} For hyper-parameters to pre-train the \textsc{Melt} and baselines, 
we pre-trained all models for 100K steps with a batch size of 128, a sequence length of 128, and a maximum learning rate of 1e-4. We utilize the linear warm-up ratio 0.048, following the previous work \cite{matscibert}. Since the materials entities have different token lengths, we adjust the mask ratios dynamically for each curriculum stage to mask 15\% tokens of the token tokens. 

\paragraph{Generation Tasks.} For generation tasks, we follow the settings in MatSci-NLP \cite{matscinlp_acl23}. Specifically, we use a learning rate of 2e-5, Adam optimizer, batch size of 4, and a maximum number of 20 training epochs with early stopping for MatSci-NLP. We also utilize 3 layers and 8 heads for decoder transformers.

\paragraph{Classification Tasks.} For classification tasks, we follow the settings in MatSciBERT \cite{matscibert}. Specifically, we use the number of epochs 20, 40, 15, and 10 for SOFC-NER, SOFC-Filling, MatScholar, and Glass Science, respectively. For each task, we sweep the learning rate in \{2e-5, 3e-5, 5e-5\}. We use Adam optimizer, and the batch size is 16.

\subsection{Evaluation} \label{appendix:evaluation}

We follow the evaluation metrics used in MatSci-NLP \cite{matscinlp_acl23} and MatSciBERT \cite{matscibert} for generation and classification tasks, respectively. For the generation benchmark MatSci-NLP, we evaluate each baseline based on both Micro-F1 and Macro-F1. We evaluate each baseline by leveraging both Micro-F1 and Macro-F1 with five different random seeds.
For classification, we evaluate each baseline using the Macro-F1 score over entity tags on SOFC-NER and SOFC-Filling, the Micro-F1 score for MatScholar, and the accuracy score for Glass Science tasks. We report the average cross-validation results over five different folds with three random seeds for classification. 


\begin{table*}[h]
\centering
\caption{The detailed ablation results for each downstream task on MatSci-NLP. CEL and MEM indicate curriculum-based entity learning and materials-aware entity masking, respectively.}
\label{tab:detailed_ablation}
{%
\begin{adjustbox}{width=\linewidth}
\begin{tabular}{cccccccccc}
\toprule
\multirow{2}{*}{Method}                                 & \multirow{2}{*}{Metric}       & \multicolumn{7}{c}{MatSci-NLP}          &                                                      \\  

 &  & NER & RC & EAE & PC & SAR & SC & SF & Overall \\ \midrule
\cellcolor[gray]{.9}  & \cellcolor[gray]{.9}Micro-F1 & \cellcolor[gray]{.9}\textbf{0.786} & \cellcolor[gray]{.9}\textbf{0.860} & \cellcolor[gray]{.9}\textbf{0.498} & \cellcolor[gray]{.9}0.728 & \cellcolor[gray]{.9}\textbf{0.798} & \cellcolor[gray]{.9}\textbf{0.911} & \cellcolor[gray]{.9}\textbf{0.610} & \cellcolor[gray]{.9}\textbf{0.741} \\

\cellcolor[gray]{.9}\multirow{-2}{*}{\textsc{Melt} (ours)} & \cellcolor[gray]{.9}Macro-F1 & \cellcolor[gray]{.9}\textbf{0.593} & \cellcolor[gray]{.9}\textbf{0.620} & \cellcolor[gray]{.9}\textbf{0.341} & \cellcolor[gray]{.9}0.647 & \cellcolor[gray]{.9}0.685 & \cellcolor[gray]{.9}\textbf{0.613} & \cellcolor[gray]{.9}\textbf{0.395} & \cellcolor[gray]{.9}\textbf{0.556} \\ \midrule 
 
 \multirow{2}{*}{\begin{tabular}[c]{@{}c@{}} w/o CEL \\ \end{tabular}} & Micro-F1 & 0.773 & 0.830 & 0.465 & \textbf{0.771} & 0.778 & 0.899 & 0.588 & 0.729 \\
 & Macro-F1 & 0.565 & 0.588 & 0.287 & 0.683 & \textbf{0.687} & 0.612 & 0.373 & 0.542 \\ \midrule
 
  \multirow{2}{*}{\begin{tabular}[c]{@{}c@{}} w/o CEL, MEM \\ \end{tabular}} & Micro-F1 & 0.728 & 0.805 & 0.495 & 0.765 & 0.739 & 0.897 & 0.559 & 0.713 \\
 & Macro-F1 & 0.528 & 0.567 & 0.311 & \textbf{0.699} & 0.621 & 0.581 & 0.303 & 0.516 \\ 
\bottomrule
\end{tabular}%
\end{adjustbox}
}
\end{table*}

\section{Curriculum Learning Baselines} \label{appendix:curriculum}
We compare our models with four different curriculum-based learning methods: frequency, masking ratio, and concept-based. We also detail the implementation of baselines. For a fair comparison, all curriculum learning baselines include a warm-up stage. We utilize 3 curriculum stages ($K=3$) and 10K steps for each curriculum stage.
\paragraph{Frequency.} We leverage the frequency of materials entities in the training corpus for defining a difficulty, similar to the previous work \cite{rare_curriculum}. We initially train the model with the most frequent materials entities and then progressively add the less frequent entities.
\paragraph{Concept.} We utilize the concept-based curriculum learning following the previous work \cite{CCM_emnlp2022}. We first extract the base concepts using both frequency and node degree and then progressively expand the entities that are related to base concepts.
\paragraph{Masking ratio.} We utilize the masking ratio as a difficulty metric, which is used as baselines in previous work \cite{CCM_emnlp2022}. Specifically, we only mask 10\% of the first sequence, and we gradually increase the masking ratio linearly to 20\% of tokens when 100K is reached.
\paragraph{Reverse.} We train the model with the reverse order of curriculum in \textsc{Melt}. Specifically, we first train with the warmup stage and then progressively train the model from stage 3 to stage 1.




\section{Detailed Ablation Results} \label{appendix:ablation}
The detailed results are shown in Table \ref{tab:detailed_ablation}. We observe that omitting each component in \textsc{Melt} leads to significant performance drops in most downstream tasks.

\newpage

\section{The Number of Curriculum Stage} \label{appendix:curriculum_num}

\begin{table}[h]
\centering
\caption{Comparison of numbers of curriculum stages on four different classification tasks.}
\begin{adjustbox}{width=\linewidth}
\begin{tabular}{@{}ccccccccc@{}}
\toprule
Methods & NER$_{\text{SOFC}}$ & NER$_{\text{MS}}$ & PC & SF & Overall \\ \midrule
2 stage  & \textbf{81.3}  & \textbf{86.0}   & 95.4  & 60.2 & 80.7     \\ 
3 stage   & 81.1   & \textbf{86.0}  & 95.7  & \textbf{62.9} & \textbf{81.4}      \\
4 stage  & 81.0 & 85.9 & \textbf{96.4} & 61.9 & 81.3       \\
\bottomrule
\end{tabular}
\end{adjustbox}
\label{table:curriculumnum}
\end{table}
To verify the design choice of the number of curriculum stages for \textsc{Melt}, we compare the performance among the model with different numbers of curriculum stages. 
The results are shown in Table \ref{table:curriculumnum}. Experimental results show that 3 stage curriculum appears to be best. Based on these results, we use 3 stages ($K=3$) in all our models.

\section{Extracted Materials Knowledge} \label{appendix:extracted_example}

\begin{figure}[h]
\centering
  \includegraphics[width=\linewidth]{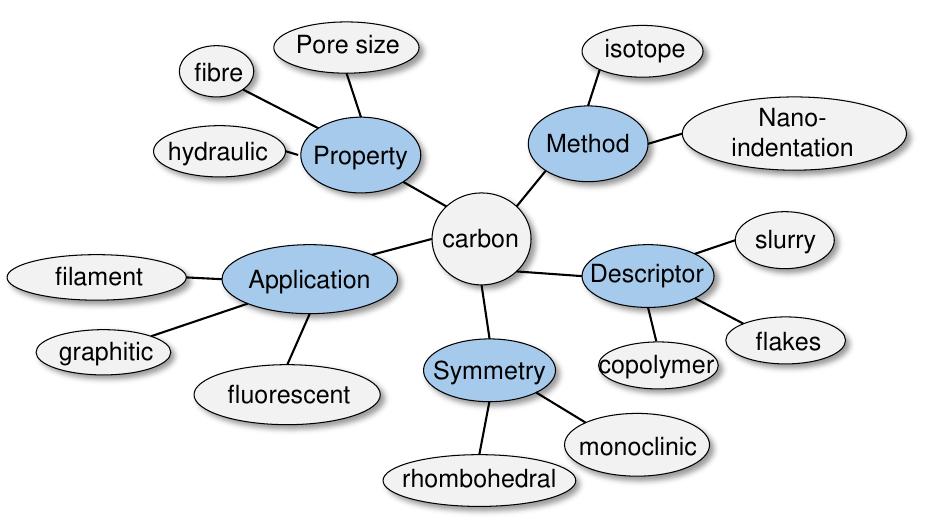}%
\caption{Examples of the constructed graphs starting from the two different chemical entities-Chemical names (\textit{carbon}). Only the entities in white circles are used to construct the set of masking words $\mathcal{G}$.}
\label{fig:analysis_graph1}
\end{figure}

\end{document}